\documentclass[runningheads]{llncs}
\usepackage{accv}
\usepackage{accvabbrv}
\usepackage{graphicx}
\usepackage{booktabs}
\usepackage[accsupp]{axessibility}  
\usepackage{comment}
\usepackage{color}
\usepackage{subcaption}
\usepackage{threeparttable}
\usepackage{adjustbox}
\usepackage{multirow}
\usepackage{multirow}
\usepackage{booktabs}
\usepackage{wrapfig,lipsum,booktabs}
\usepackage{hhline}
\usepackage{siunitx, array} 
\usepackage{wrapfig}
\usepackage{pdfpages}
\usepackage{float}
\usepackage{hyperref}
\usepackage{orcidlink}

\begin{document}
\def\methodname{Dessie}
\def\dinohmr{DinoHMR}
\def\dloss{$L_{DFL}$}
\def\pipeline{DessiePIPE}

\title{Dessie: Disentanglement for Articulated 3D Horse Shape and Pose Estimation from Images} 

\titlerunning{Dessie}

\author{Ci Li\inst{1}\orcidlink{0000-0002-7627-0125} \and
Yi Yang\inst{1,2}\orcidlink{0000-0002-6679-4021} \and
Zehang Weng\inst{1}\orcidlink{0000-0002-9486-9238} \and \\
Elin Hernlund\inst{3}\orcidlink{0000-0002-5769-3958} \and 
Silvia Zuffi\inst{4}\orcidlink{0000-0003-1358-0828} \and
Hedvig Kjellström\inst{1,3}\orcidlink{0000-0002-5750-9655} 
}

\authorrunning{C.~Li et al.}

\institute{
KTH, Sweden \email{\{cil, yiya, zehang, hedvig\}@kth.se}\\
\and 
Scania, Sweden \email{carol-yi.yang@scania.com} \\
\and
SLU, Sweden \email{Elin.Hernlund@slu.se} \\ 
\and 
IMATI-CNR, Italy \email{silvia@mi.imati.cnr.it}
}

\maketitle

\begin{figure}[h]
    \centering
    \includegraphics[width=\textwidth]{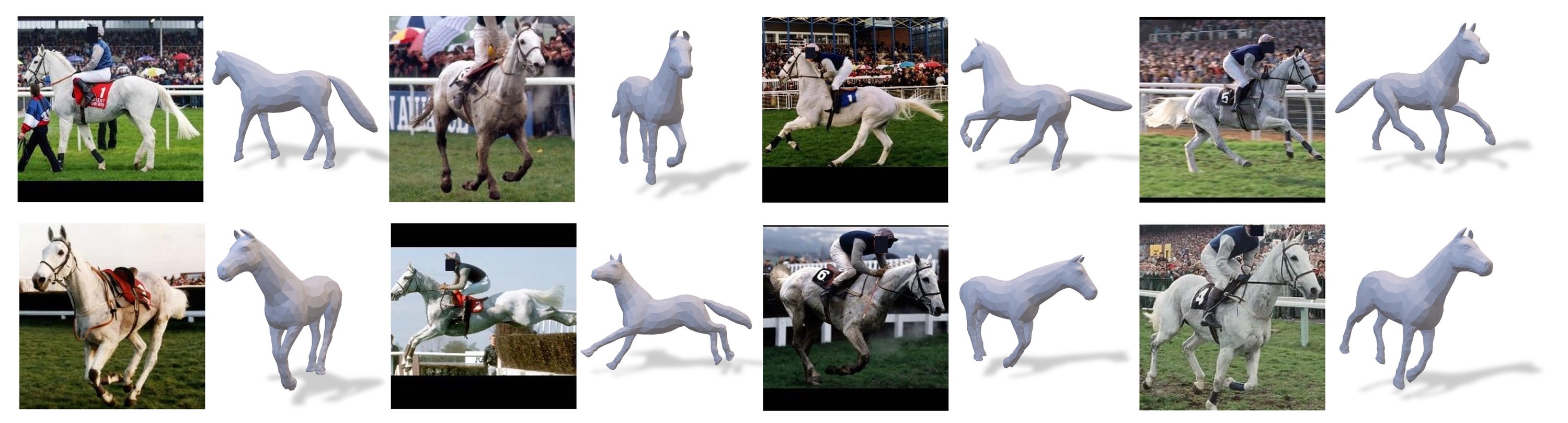}
    \caption{We estimate 3D shape and pose of horses from monocular images. The figure shows pictures of Dessie, a famous racehorse, together with our 3D reconstruction. }
    \label{fig:front}
\end{figure}
\begin{abstract}

In recent years, 3D parametric animal models have been developed to aid in estimating 3D shape and pose from images and video. While progress has been made for humans, it's more challenging for animals due to limited annotated data. To address this, we introduce the first method using synthetic data generation and disentanglement to learn to regress 3D shape and pose. Focusing on horses, we use text-based texture generation and a synthetic data pipeline to create varied shapes, poses, and appearances, learning disentangled spaces. Our method, Dessie, surpasses existing 3D horse reconstruction methods and generalizes to other large animals like zebras, cows, and deer. See the project website at: \url{https://celiali.github.io/Dessie/}.

\end{abstract}
\section{Introduction}
\label{sec:intro}

Capturing 3D articulated motion of humans and animals is essential for understanding motion dynamics and enhancing various applications such as virtual reality~\cite{10.1145/3528223.3530086}. 
In addition, motion can be often associated with diseases and pain, making motion capture an effective diagnosis tool for animal subjects that do not communicate discomfort verbally. Recent advances in computer vision have promoted 3D markerless motion capture~\cite{tian2023recovering}, allowing motion inference from visual data without invasive equipment. 
Substantial efforts in these markerless approaches have focused on using parameterized models like SMPL for humans~\cite{loper2015smpl} and SMAL for animals~\cite{zuffi20173d}, which define local (shape and pose) and global (rotation and translation) properties. Their precise capture makes them important analysis tools across various fields, particularly in biomedical downstream tasks~\cite{keller2023skel,Keller:CVPR:2024,LUO202336}.

In this work, we focus on horses. Horses are particularly relevant for their economic value~\cite{AmericanHorseCouncil2024,IBISWorld2024}. Maintaining horses requires substantial resources~\cite{TransparencyMarketResearch2024}. Moreover, despite their large bodies, horses are delicate, owning an anatomical fragility that often leads to severe, irrecoverable injuries, in particular for the most precious racing subjects. Consequently, horses have been extensively studied from both behavioral and biomechanics perspectives. In computer vision, recent efforts have developed 3D articulated parameterized shape models specific for horses~\cite{li2021hsmal,Zuffi:CVPR:2024}, confirming our view that horses, for their particular wide presence in several human activities, require ad-hoc, specific models.
What the horses have in common with other animals is the scarcity of 3D annotated datasets that would facilitate learning to regress 3D poses from images with end-to-end regressors. In addition, horses might share similarities in how they move with other species of the same Equine family. Motivated by these observations, we explore disentanglement in learning image-to-3D model parameter regressors.

Disentanglement is a powerful machine learning technique, offering advancements in different fields like image generation~\cite{higgins2016beta} and style transfer~\cite{kotovenko2019content}. By separating underlying factors of variation in the data, it facilitates more interpretable models and enhances generalization and robustness. In image recognition, disentangled representations have achieved state-of-the-art (SOTA) results~\cite{qing:2023:disentangling,hao:2023:learning}, showcasing their ability to isolate and manipulate salient features in complex data. Motivated by these successes, we investigate the potential of disentanglement to address challenges in monocular 3D horse mesh reconstruction.

Recent advancements in 3D markerless motion capture domain have shown that regression-based approaches, utilizing neural networks to directly map image pixels to model parameters, achieve SOTA results for human~\cite{kanazawa2018end,humanMotionKanazawa19,kocabas2020vibe,goel2023humans} and animal~\cite{zuffi2019three,biggs2020left} modeling. These methods employ encoders~\cite{he2016deep,dosovitskiy2020image}, for feature extraction, with decoders predicting model parameters from the extracted features.
However, these methods embed all image attributes—ranging from subject shape and pose to the background—into a single feature space, entangling diverse properties, and complicating the decoder's task of isolating relevant information. 
In this work, we employ a specialized encoder to separate features relevant 
to the subject's local (pose, shape) and global information (rotation, translation, camera), coupled with distinct decoders to regress separate model parameters. To the best of our knowledge, this is the first work in which disentangled feature spaces are used for animal 3D reconstruction. 

To enable disentanglement learning and regression-based 3D reconstruction of animals, a large-scale dataset is essential. Collecting real-world data is challenging, primarily due to the scarcity of real-world animal data, and the high costs associated with data collection, unlike the relative abundance of available human data~\cite{goel2023humans}. 
To address this challenge, we develop a synthetic pipeline, \pipeline{}, capable of generating realistic animal images on the fly, to support regression-based 3D animal reconstruction while encouraging disentanglement learning. \pipeline{} utilizes a newly published horse motion capture dataset~\cite{PFERD} and the hSMAL model~\cite{li2021hsmal}, rendering horse images, in various poses (standing, walking, trotting, cantering, eating, bending neck, sitting, rearing), shapes, and species-related textures, with randomized orientations and backgrounds. Additionally, we introduce a novel technique, to create image tuples with controllable variations, allowing the study of disentangled features by producing pairs of images that vary one feature while maintaining others (\eg different appearances with the same pose and global rotation).

Leveraging \pipeline{}, we present \dinohmr{}, a variant of the popular HMR framework~\cite{kanazawa2018end}, and introduce~\methodname{}\footnote{Desert Orchid (1979–2006), known as Dessie, was one of the most beloved and iconic racehorses in British history. Source: Wikipedia.}, a multi-stream framework combining disentanglement learning. Both methods are based on the vision foundation model DINO~\cite{caron2021emerging}. Unlike HMR, \methodname{} disentangles image features into different spaces with independent processing paths, such that information that pertains to disentangled features can propagate through the layers without spurious influences. 

Through extensive experiments, we validate our methods, demonstrating their effectiveness by training exclusively on synthetic data from \pipeline{}. Our methods generalize well across unseen real-world datasets and deliver both qualitative and quantitative improvements when fine-tuned on real image datasets of varying sizes, from limited to large. \methodname{} outperforms existing 3D animal reconstruction methods even with limited real data, offering a solution to data scarcity, and generalizes well across out-of-distribution image domains and to other horse-like species. 

To summarize, our contributions are:
\begin{itemize}  
\item We propose \pipeline{}, a method to generate diverse and high-quality synthetic images for 3D horse markerless reconstruction learning.
\item We present \methodname{}, a framework for 3D pose and shape estimation of horses from monocular images based on the foundation model DINO; on disentangled latent spaces; and on disentangled learning. For comparison, we also propose \dinohmr{} (\methodname{} without disentanglement).
\item We show that disentangled learning results in more efficient training, and achieves SOTA performance on various benchmarks.
\end{itemize}

\section{Related Work}
\label{sec:relatedwork}

\paragraph{Model-based methods for 3D reconstruction of humans and animals.}
The markerless motion capture of articulated deformable subjects, like humans and many animals, requires prior models of body shape and pose when applied in monocular settings. In recent years, the development of 3D parametric shape models of the human body has seen explosive development in computer vision and graphics, with the SMPL model \cite{loper2015smpl} being the de-facto standard.
SMPL and its successors~\cite{loper2015smpl,pavlakos2019expressive,osman2020star,osman2022supr} are data-driven models learned from thousands of 3D body scans of real people. 
They define statistical parametric linear shape models, for generating bodies of different shapes, sizes, and genders.  
A huge number of methods~\cite{bogo2016keep,kanazawa2018end,8099983,8578153,kolotouros2019learning,humanMotionKanazawa19,kocabas2020vibe,9577652,9710873,9711119,10.1007/978-3-031-20065-6_34,9879793,li2023niki} focus on recovering 3D shape and pose from monocular images or videos using SMPL models.   

Given the challenges of learning animal models from 3D captured data, initial research on animal modeling focused on creating 3D models from images. A pioneering effort by Cashman and Fitzgibbon demonstrates the reconstruction of dolphins and bears from images~\cite{cashman2012shape}. 
Later, Zuffi et al. introduce SMAL~\cite{zuffi20173d}, the animal equivalent of SMPL, learned from 3D scans of a limited set of toys representing different quadrupeds. 
Based on the SMAL model, Zuffi et al.~\cite{zuffi2019three} introduce an end-to-end regression network to reconstruct zebras, marking an important step towards more detailed and accurate animal 3D model-based monocular reconstruction.
The 3D capture of dogs has seen innovation, with Kearney et al.~\cite{kearney2020rgbd} utilizing RGBD images for reconstruction, while Biggs et al.~\cite{biggs2020left,biggs2019creatures} achieve pose and shape estimation from solely image data. Rüegg et al.~\cite{ruegg2022barc,ruegg2023bite} enhance accuracy by incorporating breed-specific information and ground contact data. Bird modeling has also progressed, employing images and a 3D synthetic model~\cite{badger20203d,wang2021birds} to learn shape variation of birds.
Li et al.~\cite{li2021hsmal} introduce the hSMAL model, a horse-specific SMAL model that facilitates the recovery of 3D poses from images and aids in lameness detection in horses~\cite{liautard1888lameness}, showing the potential of these technologies for biological and medical applications. VAREN~\cite{Zuffi:CVPR:2024} builds upon hSMAL, leveraging real horse scan data.

Many existing regression methods~\cite{kanazawa2018end,kolotouros2019learning,kocabas2020vibe,li2023niki,10.1007/978-3-031-20065-6_34,zuffi2019three,biggs2020left} encode images as a single feature vector to directly predict camera, model shape and pose parameters.
Notably, some methods take a different approach.
PARE~\cite{9711119} employs part-guided attention from one module alongside 3D features from another to better understand occluded body parts.
HybrIK~\cite{9577652} leverages separate modules to generate 3D heatmaps and predict shape and twist angles for inverse kinematics.
Pavlakos et al.~\cite{8578153} estimate silhouette and heatmap from images to infer model shape and pose, showing that disentangling shape and pose enhances 3D prediction stability and accuracy. Recent works on dogs, Barc~\cite{ruegg2022barc} and Bite~\cite{ruegg2023bite}, employ separate pose and shape modules.
Unlike these approaches, our work proposes to learn three separate feature spaces for shape, pose, and global information (rotation, translation and camera) prediction in a disentangled fashion. 

\paragraph{Feature Extraction Networks.}
The backbone networks for extracting image features are diverse, with ResNet~\cite{he2016deep}, Stack Hourglass\cite{newell2016stacked}, and HRNet~\cite{sun2019deep} being commonly used. 
Recently, DINO~\cite{caron2021emerging}, a self-supervised self-distillation large-scale foundation model, has proven effective in extracting image representation~\cite{tumanyan2022splicing} and performs well on downstream tasks~\cite{amir2021deep, tschernezki2022neural}. Pretrained DINO features have been used in non-template based 
3D animal reconstruction tasks, like, in Lassie~\cite{yao2022lassie}, Hi-Lassie~\cite{yao2023hi}, and Artic3D~\cite{yao2024artic3d} to optimize 3D animal shape from a small scale of images (less than 30 images); in Magicpony~\cite{wu2023magicpony}, Farm3D~\cite{jakab2023farm3d}, 3D Fauna~\cite{li2024learning} to learn 3D priors from a large-scale dataset (over 10k images).
Similar to our work, Digidogs~\cite{Shooter_2024_WACV} fine-tunes DINOv2~\cite{oquab2023dinov2} for estimating 3D dog skeletons of real-world images using simulation data by extracting token features from DINOv2. Ours more focus on the extracted key features from DINO.

\paragraph{Disentanglement.} 
Disentanglement aims to figure out the underlying latent representations that control different data properties. It has be applied to both 2D images~\cite{chen2016infogan, higgins2016beta,lorenz2019unsupervised} and 3D meshes~\cite{zhou2020unsupervised}.
Lorenz et al.~\cite{lorenz2019unsupervised} disentangle shape and appearance from 2D images by learning parts consistency over category instances, leveraging the equivariance and invariance constraints on the image representation to extract part-based information.
Zhou et al.~\cite{zhou2020unsupervised} disentangle shape and pose features from registered meshes by using self-consistency and cross-consistency constraints in an unsupervised setting. 
Hu et al.~\cite{hu2022disentangling} extract disentangled 3D attributes only from 2D images to manipulate the 3D output, based on PiFu, which decodes image features from the backbone network into 3D implicit representations of humans. The image features encode three attributes, which are disentangled by the designed frameworks.
Zhao et al.~\cite{zhao2021learning} disentangle pose and view factors from 2D human poses by maximizing mutual information of the same pose from different views in a contrastive learning manner.
While these works focus on 2D pose estimation, 3D mesh generation or 3D mesh manipulation tasks, none address monocular model-based 3D reconstruction. 

To encourage disentanglement, contrastive learning is an effective strategy. Traditional contrastive methods, such as SimCLR~\cite{chen2020simple} and MoCo~\cite{he2020momentum}, typically involve pre-training an image encoder on large datasets in an unsupervised manner. These methods gain their effectiveness by generating positive and negative image pairs, which encourages the model to closely align the features of the positive pairs. Our method simplifies the process by leveraging the labels from the generated image pairs and ensuring feature similarity for each matching pair.
\section{Method}
\label{sec:method}

\subsection{The hSMAL model} 
Analogous to the SMPL model, the hSMAL model~\cite{li2021hsmal}, a horse-specific version of the SMAL model, is a parameterized 3D model of horses. It factorizes surface variations into shape, and pose. Given 37 horse toy scans as training data, the average shape over the training set is calculated as $V_{mean}$, and PCA is performed on the variations of each scan to the mean shape. The shape, noted as $\beta \in R^9$, is the set of PCA coefficients representing vertex-based deformations from the mean shape. The pose, noted as $\theta =(\theta_{G}, \theta_{J})$, represents the global rotation $\theta_{G} \in R^3$ and the relative rotation of each joint with respect to its parents in the kinematic tree $\theta_{J} \in R^{35 \times 3}$ in axis angle representation. The hSMAL model defines a mapping function, $\Theta : (\beta, \theta, \xi) \mapsto \mathbf{v}$, which outputs a 3D mesh based on linear blend skinning.  $\xi$ is the global translation.

\subsection{\pipeline{}}
As noted, to advance disentanglement learning and regression-based 3D reconstruction techniques, we require a large-scale animal dataset. However, compiling this dataset through real-world data collection presents significant challenges. Particularly, it's highly impractical to obtain e.g. a pair of animal images in identical poses with different appearances, which is crucial for promoting feature disentanglement by having image pairs that share similar features with variations in another feature.
\begin{figure}[t]
    \centering
\includegraphics[width=0.85\linewidth]{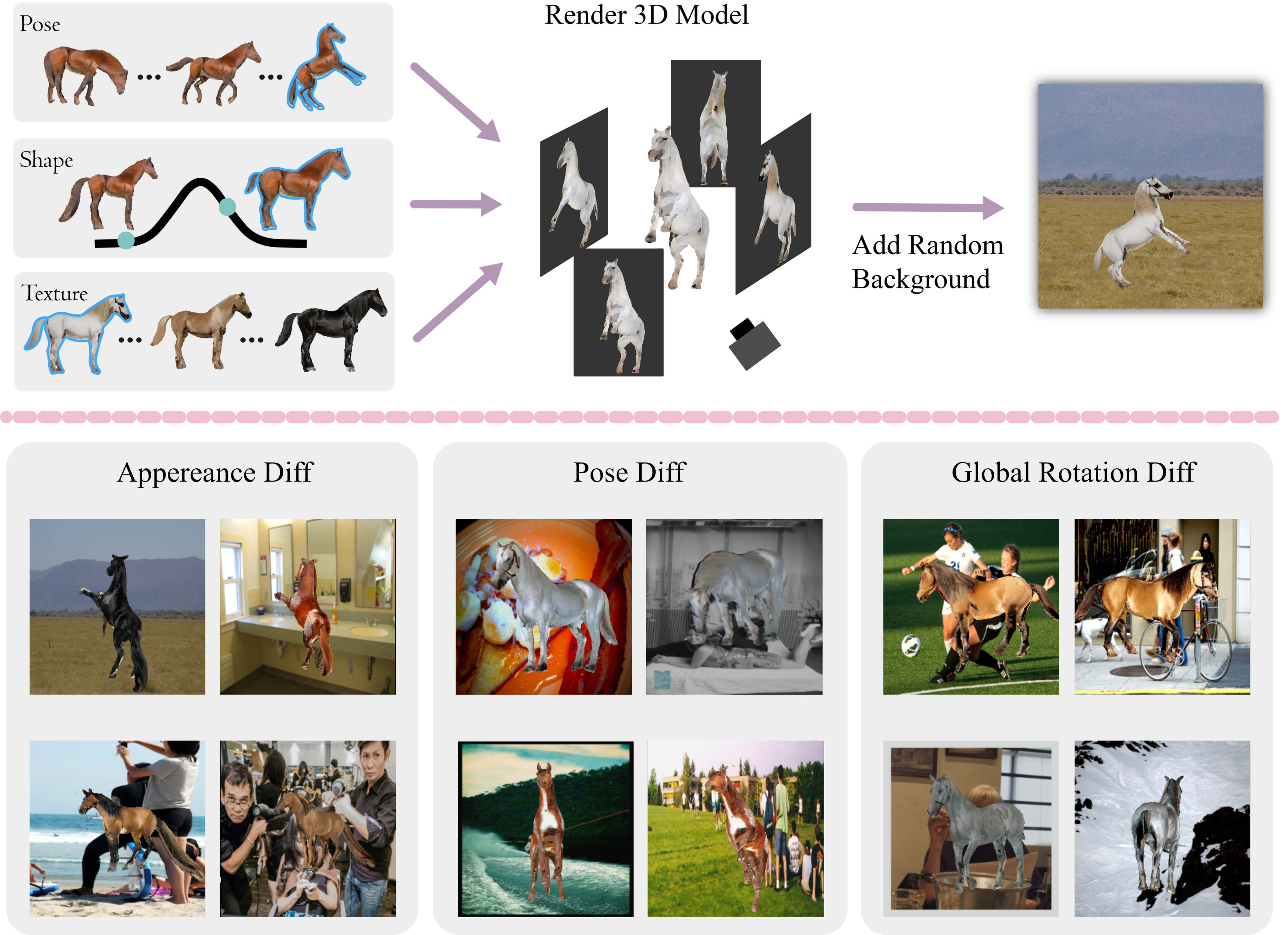} 
    \caption{\pipeline{}: Synthetic generation pipeline. Top: the training image generation process; bottom: generated data samples with controlled variations.}
    \label{fig:data_pipeline}
\end{figure}
To overcome this challenge, we develop \pipeline{}. \pipeline{} allows for the creation of horse images with diverse combinations of camera viewpoints, 
horse appearances, and poses, utilizing Pytorch3D~\cite{ravi2020pytorch3d} and the hSMAL model as shown in Fig~\ref{fig:data_pipeline}. It differs from previous offline synthetic data generation, using limited textures and poses learned from 2D images\cite{xu2023animal3d}, or fixed shape \cite{mu2020learning,li2021synthetic}. A concurrent method is FoundationPose~\cite{wen:2024:foundationpose}, which generates synthetic data using language and diffusion models for RGBD 3D rendering.

To generate realistic horse appearances, 
we create a finite set containing 80 highly realistic UV texture maps following TEXTure~\cite{richardson2023texture}, which generates texture maps for a given 3D shape with a text prompt by leveraging pretrained depth-to-image stable diffusion model~\cite{rombach2022high}. Specifically, the texture set is constructed with text prompts based on eight distinct horse species, including Bay Thoroughbred, Palomino Quarter Horse, Chestnut Morgan, Buckskin Tennessee Walker, White Arabian, Black Friesian, Dapple Gray Andalusian, Pinto Paint Horse.
We refer the readers to \textit{Supplementary Materials} for more details. For the horse's shape, we obtain an infinite set by randomly sampling within the hSMAL shape space with a Gaussian distribution. Regarding the pose, we create a finite set composed of realistic horse poses extracted from PFERD~\cite{PFERD}, including daily motion poses (standing, walking, trotting, eating, bending neck) and advanced poses (sitting, rearing). Note that these poses also encapsulate the global rotations. 
Random image backgrounds come from the COCO dataset~\cite{lin2014microsoft}.
With the described components, \pipeline{} randomly selects an item from each of the appearance, pose, and background sets to compose a horse image. This process is further exploited by \pipeline{} to generate pairs of images $(I1, I2)$, where $I1$ and $I2$ are images that differ in one aspect—either model shape, pose, or global rotation. Fig~\ref{fig:data_pipeline} provides examples of how individual changes impact the rendered outcomes.
Images are generated using Pytorch3D, each with a resolution of (256, 256). Ground-truth annotations for each image include: global rotation $\theta_{G_{gt}}$, pose $\theta_{J_{gt}}$, shape $\beta_{gt}$, global translation $\xi_{gt}$, silhouette $S_{gt}$, and landmark locations $K_{gt}$. We define 17 landmarks on the 3D model template, corresponding to the 2D animal keypoint labels in ViTPose+~\cite{xu2022vitpose+}, and apply the Pytorch3D rasterizer to filter out and retain landmarks visible in the camera views.

\subsection{Network Architecture}

\begin{figure}[t]
    \centering
    \includegraphics[width=0.93\linewidth]{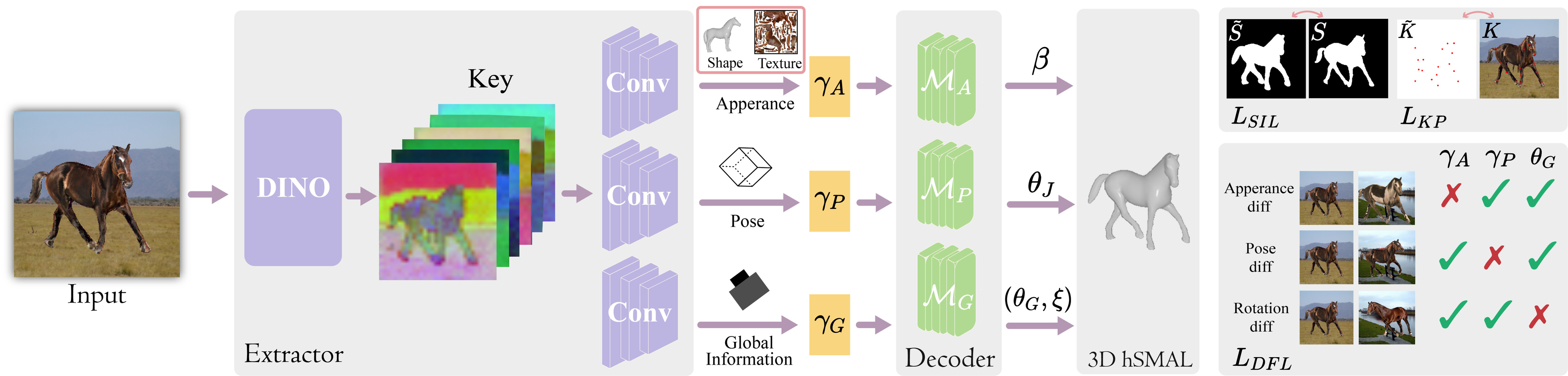}
    \caption{Network architectures of \methodname{}. \methodname{} extracts latent features ($\gamma_A$, $\gamma_P$, $\gamma_G$) with DINO and predicts the hSMAL parameters with the corresponding decoders. The model is trained with keypoint loss, silhouette loss and a contrastive loss to encourage the disentanglement.}
    \label{fig:network_architecture}
\end{figure}

Inspired by HMR~\cite{kanazawa2018end} and DINO-related feature studies~\cite{caron2021emerging, tumanyan2022splicing, amir2021deep, wu2023magicpony}, we propose two extractor-decoder frameworks that incorporate the DINO model: 1) \dinohmr{}, an HMR-like model,  
and 2) \methodname{}, a multi-stream structure with a modified contrastive loss \dloss{}, 
for disentangled shape and pose estimation (Fig~\ref{fig:network_architecture}).
Unlike HMR, which encodes 
image features with ResNet, both \dinohmr{} and \methodname{} exploit the capabilities of DINO's key features from the last layer. 

In \dinohmr{}, we employ an \textit{extractor} with a Conv2D network to extract a single feature representation from the keys, and a \textit{decoder} for predicting shape $\beta$, pose $\theta_J$, global rotation $\theta_G$ and translation $\xi$ parameters.
In contrast, the \methodname{} \textit{extractor} $\mathcal{F}(.)$ employs three Conv2D networks on the keys to isolate features $\gamma_{A} \in \Gamma_{A}$, $\gamma_{P} \in \Gamma_{P}$, and $\gamma_{G} \in \Gamma_{G}$, corresponding to the feature spaces for subject appearance, subject pose, and global information, respectively, with $\Gamma_{A}, \Gamma_{P}, \Gamma_{G} \subseteq \mathbb{R}^{640}$.
The \methodname{} \textit{decoder} is composed of three distinct submodules $\mathcal{M}_A$, $\mathcal{M}_P$, and $\mathcal{M}_G$, each processing a specific set of features extracted by the \methodname{} \textit{extractor}. As shown in Fig.~\ref{fig:network_architecture}, the submodule $\mathcal{M}_A$ is responsible for inferring the shape parameters $\beta$ from the appearance features $\gamma_A$. The submodule $\mathcal{M}_P$ deduces the pose parameters, $\theta_J$ leveraging the pose features 
$\gamma_P$. Lastly, $\mathcal{M}_G$ estimates both the global rotation $\theta_G$ and the global translation $\xi$, utilizing the global information features $\gamma_G$. The entire framework is described by:

\begin{small}
\begin{equation}
\begin{aligned}
\text{\methodname{} Extractor:} & \quad (\gamma_{A}, \gamma_{P}, \gamma_{C}) = \mathcal{F}(x), \\
\text{\methodname{} Decoder:} & \quad \beta = \mathcal{M}_A(\gamma_{A}), \\
& \quad \theta_{J} = \mathcal{M}_P(\gamma_{P}), \\
& \quad (\theta_{G}, \xi) = \mathcal{M}_G(\gamma_{G})
\end{aligned}
\label{eq:feature_embed}
\end{equation}
\end{small}

We employ a weak-perspective camera model $\Omega (s, t_x, t_y)$, with learnable scale $s \in R $ and translation $(t_x, t_y) \in R^2 $, and pre-defined focal length $f = 5000$. Given the relativity between the camera and the 3D model, the global translation of the model is defined as $\xi = \left[ t_x, t_y, \frac{2 \cdot f}{r \cdot s } \right]$, where $r = 256 $ is the image resolution. We use a differentiable render, Pytorch3d, to render the silhouette $\tilde{S}$ and 2D keypoints $\tilde{K}$ by projecting the model back to the image. 

\subsection{Loss}\label{sec:method_loss}

\methodname{} is trained in an end-to-end manner, with loss defined as: 
\begin{small}
\begin{equation}
L = L_{KP} + L_{SIL} + L_{HSMAL} + L_{DFL}.
\label{eq:total_loss_function}
\end{equation}
\end{small}
$L_{KP}$ is a 2D keypoint loss defined as :
\begin{small}
\begin{equation}
L_{KP} = \omega_{K} \frac{ \sum_{j=1}^{J} \lambda_{j}^2 \rho \left\| \tilde{K}^{{j}} - K_{gt}^{{j}} \right\|_{2} }{\sum_{j=1}^{J} \lambda_{j}^2}, \ \tilde{K}^{{j}} =\Pi(\kappa^{j}, \Omega),
\end{equation}
\end{small}
where $\kappa$ are the body keypoints on the model, projected by the camera $\Omega$. $K_{gt}$ are the ground-truth 2D keypoints with visibility flag $\lambda$, and $\rho$ is the Geman-McClure robustifier~\cite{geman1987statistical}. $\Pi$ is the differentiable rendering function.
$L_{SIL}$ is a silhouette loss defined as:
\begin{small}
\begin{equation}
L_{SIL} = \omega_{S} smoothL1( \tilde{S} , S_{gt} ), \ \tilde{S}=\Pi ( \mathbf{v} , \Omega),
\end{equation}
\end{small}
where $S_{gt}$ is the groundtruth silhouette, $\tilde{S}$ is the projected model silhouette, and $smoothL1(.)$ is the SmoothL1 loss~\cite{girshick:2015:fastrcnn}. $L_{HSMAL}$ is the weighted sum of the shape and the pose priors of the hSMAL model, defined in \cite{li2021hsmal}, with corresponding weights \begin{small}$\omega^{\beta}_{Prior}$\end{small} and \begin{small}$\omega^{\theta}_{Prior}$\end{small}.
\dloss{} is a variant of contrastive loss designed to encourage the disentanglement between different features and minimize intra-pair differences with known input pair variations, defined as:
\begin{small}
\begin{equation}
L_{DFL} = \omega_D \begin{cases}
MSE(\gamma_{P_1}, \gamma_{P_2}) + MSE(\theta_{G_1}, \theta_{G_2}), & \text{ if appearance difference.} \\ 
MSE(\gamma_{A_1}, \gamma_{A_2}) + MSE(\theta_{G_1}, \theta_{G_2}), & \text{ if pose difference.} \\ 
MSE(\gamma_{A_1}, \gamma_{A_2}) + MSE(\gamma_{P_1}, \gamma_{P_2}), & \text{ if global rotation difference.}
\end{cases}
\label{eq:DFL_loss_function}
\end{equation}
\end{small}

Given the ambiguity between the camera and model, the loss on global translation is not applied. 
It is important to note that ground-truth shape and pose labels from \pipeline{} are not necessary for the learning process. For \dinohmr{}, the contrastive loss term \dloss{} is omitted. 
\section{Experiments}
\label{sec:exp}

\subsection{Datasets}

\paragraph{Synthetic Dataset.}
Through \pipeline{}, we construct synthetic validation dataset, using 10\% of the texture maps and around 12\% of the poses, and the validation set from the COCO dataset for the background. For each training epoch, \pipeline{} creates 6.4k and 640 images for training and validation.

\paragraph{Real-world Dataset.}
Our evaluation utilizes several public datasets:
1) The horse category from PASCAL~\cite{everingham2015pascal} and AnimalPose~\cite{cao2019cross};
2) The horse category in MagicPony~\cite{wu2023magicpony} dataset, which compiles images from Weizmann Horse Database~\cite{borenstein2002class}, PASCAL~\cite{everingham2015pascal}, and Horse-10 Dataset~\cite{mathis2021pretraining}.
Each real-world dataset includes annotated segmentation masks. Given the keypoints labels are not available in the MagicPony dataset, we generate keypoint labels with ViTPose+~\cite{xu2022vitpose+}.

\subsection{Network Implementation Details}

\dinohmr{} and \methodname{} both use DINO-ViTs8 as the backbone for the extractor $\mathcal{F}$, with the last three layers unfrozen for adaptation to the 3D animal reconstruction task.
In the decoder $\mathcal{D}$, each module comprises two fully connected layers with dropout applied in between, followed by a fully connected layer to predict the residual of each parameter. The associated features and the corresponding parameters are concatenated as input to each module, which outputs residual to update the parameters for three iterations. 
Training spans up to 800 epochs, with a learning rate of $5 \times 10^{-5}$. Model selection is based on achieving the lowest loss on the validation dataset. For the weights in the loss function, from Eq.~\ref{eq:total_loss_function} to Eq.~\ref{eq:DFL_loss_function}, we set $\omega_K = 0.001$, $\omega_S = 0.0001$, $\omega^{\beta}_{\text{Prior}} = 0.01$, $\omega^{\theta}_{\text{Prior}} = 0.01$, and $\omega_D = 0.02$. We apply color jittering to the images. The training process is conducted on two Nvidia A100 GPUs for approximately two days.

\subsection{Experiments}

We conduct two main experiments to demonstrate the effectiveness of \methodname{}. First, we train both \dinohmr{} and \methodname{} solely based on synthetic data generated by \pipeline{} with the same sampling strategy, and validate their performance on unseen real-world datasets. Second, we fine-tune \dinohmr{} and \methodname{} using either a limited amount of real data from Staths~\cite{stathopoulos2023learning} or a comparatively larger dataset from MagicPony~\cite{wu2023magicpony}, to showcase the model's capacity for generalization across different scales of data.

\paragraph{\dinohmr{} vs. \methodname{} with \pipeline{}.}

Given that \pipeline{} is fully controllable and observable, we can incorporate ground-truth shape and pose labels from \pipeline{} for supervised learning. We introduce a ground-truth loss term $L_{\text{gt}}$, for both shape $\beta$ and pose $\theta$, using the mean squared error (MSE) between the predicted values and the ground truth labels from \pipeline{}. 

\begin{table}[t]
\centering
\caption{\dinohmr{} vs. \methodname{} with \pipeline{} on unseen image datasets.}\label{tab:foundational_analysis}
\begin{tabular}{c|c|c|cc|cc|cc} 
\toprule
Network & $L_{gt}$ & \dloss{} & \multicolumn{2}{c|}{Magicpony Dataset} & \multicolumn{2}{c|}{\hspace{12px} AnimalPose \hspace{12px} } & \multicolumn{2}{c}{\hspace{15px} Pascal \hspace{15px} } \\ 
\cline{4-9}
& & & PCK\tiny{@0.1}\normalsize($\uparrow$) & IoU($\uparrow$) & PCK\tiny{@0.1}\normalsize($\uparrow$) & IoU($\uparrow$) & PCK\tiny{@0.1}\normalsize($\uparrow$) & IoU($\uparrow$) \\ 
\midrule
\multirow{2}{*}{\dinohmr{}} & - & - & 67.70 & 57.39 & 68.42 & 56.87 & 47.65 & 47.63 \\
                         & $\checkmark$ & - & 61.02 & 54.68 & 68.70 & 54.87 & 45.23 & 45.85 \\ 
\midrule
\multirow{4}{*}{\methodname{}}& - & - & 59.94 & 54.61 & 67.13 & 55.52 & 43.11 & 44.58 \\
                         & $\checkmark$ & - & 69.42 & 58.02 & 73.63 & 58.24 & 47.73 & 46.93 \\
                         & $\checkmark$ & $\checkmark$ & 65.66 & 56.39 & 70.26 & 56.15 & 45.15 & 47.69 \\
                          & - & $\checkmark$ & \textbf{70.97} & \textbf{59.99} & \textbf{75.43} & \textbf{59.69} & \textbf{52.27} & \textbf{52.51} \\
\bottomrule
\end{tabular}
\footnotesize Note: $\checkmark$ indicates inclusion of the loss component in training; - indicates exclusion.
\end{table}

We examine the performance of \dinohmr{} and \methodname{} when utilizing the contrastive loss \dloss{},
in comparison to their supervised-learning variants employing $L_{\text{gt}}$, as presented in Table~\ref{tab:foundational_analysis}. Notably, \dinohmr{} and \methodname{} are trained without any real images. 
Our evaluation metrics include the Intersection Over Union (IoU) and the Percentage of Correct Keypoint (PCK) with a threshold of 0.10, applied to manually defined keypoints on the model across all unseen real-world datasets.
For MagicPony dataset, we report PCK among the 17 predicted keypoints defined by ViTPose+. For Pascal and AnimalPose datasets, we follow the evaluation in~\cite{stathopoulos2023learning}, employing 16 keypoints as specified for Pascal.
\methodname{} (w/o $L_{gt}$, w/ \dloss) consistently outperforms other methods, supporting our hypothesis that a model promoting feature disentanglement achieves better generalization in zero-shot real image reconstructions. 

\begin{figure}[t]
    \centering
    \includegraphics[width=0.74\linewidth]{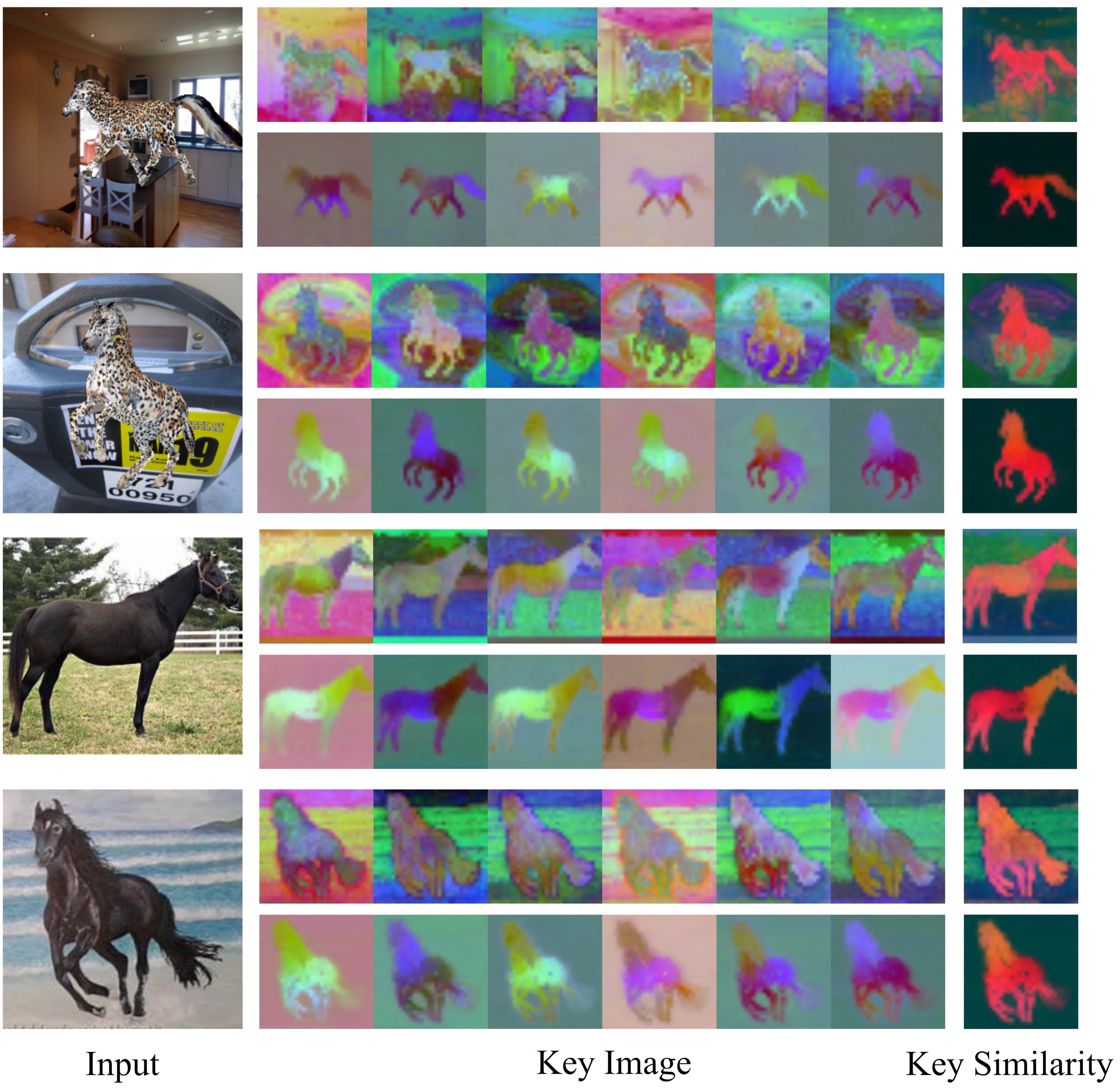}
    \caption{Visualization of the leading PCA components of key features for two synthetic and two real images from top to down. For each image, we visualize the original DINO (row 1) and the \methodname{} key features (row 2).}
    \label{fig:pca_dino}
\end{figure}

\paragraph{DINO Key Feature.}
Given that we train \methodname{} with DINO backbone unfrozen, we are interested in one question:
\textit{How does the 3D reconstruction learning affect the DINO feature space?}
To answer this question, we utilize PCA to visualize the key features extracted from the DINO model's last layer, and examine the self-similarity of these features, as in previous studies~\cite{tumanyan2022splicing}. Brighter areas represent higher component values, indicating regions of intense focus for reconstruction. 
This analysis uses two synthetic test images and two real images from the AnimalPose dataset, as shown in Fig~\ref{fig:pca_dino}. The comparison reveals that while the vanilla DINO features capture the spatial layout and structure of both the subjects and environments, the features in \methodname{} shift to focus more toward the targeted reconstructed subjects, effectively minimizing background distractions.

\paragraph{\methodname{} vs. SOTA with limited real shots.}

We assess the performance of \dinohmr{} and \methodname{} against A-CSM~\cite{kulkarni2020articulation} and Staths~\cite{stathopoulos2023learning}, with a constrained training dataset of 150 labeled images~\cite{stathopoulos2023learning}. Notably, Staths is an augmented A-CSM framework with additional web-sourced images for training.
During the training, we construct batches to maintain an equal proportion of synthetic and real images, selecting model based on the performance of the synthetic validation set. Real images are excluded from computing \dloss{} and the silhouette loss $L_{SIL}$ to maintain consistency with~\cite{stathopoulos2023learning} during training. 
The effectiveness is quantified on AnimalPose and Pascal datasets as shown in Table~\ref{tab:performance_comparison_Animal3d}, using the Area Under the Curve (AUC) metric to capture PCK performance across varying thresholds from 0.06 to 0.1. The results show that both \dinohmr{} and \methodname{}, even when fine-tuned with only the 150 images, outperform Staths, highlighting their effectiveness in utilizing limited real-world data resources.

\begin{table}[H]
    \centering
    \caption{KeyPoint Evaluation. $\dagger$: number taken from~\cite{stathopoulos2023learning}.  }
    \label{tab:performance_comparison_Animal3d}
    \begin{adjustbox}{max width=\linewidth}
    \setlength{\tabcolsep}{4pt}
        \begin{tabular}{lccccccc}
            \toprule
            AUC ($\uparrow$)& A-CSM~\cite{kulkarni2020articulation}$\dagger$ & Staths~\cite{stathopoulos2023learning}$\dagger$ & \dinohmr{} & \methodname{} & \dinohmr{}$\circledast$ & \methodname{}$\circledast$ \\
            \midrule
            AnimalPose & 51.0 & 75.1 & 55.6 & 62.2 & 81.8 & \textbf{82.9} \\
            \midrule
            Pascal & 37.4 & 55.9 & 36.4 & 40.7 & 60.5 & \textbf{61.2} \\
            \bottomrule
        \end{tabular}
    \end{adjustbox}
    \footnotesize Note: $\circledast$: Model finetuned on dataset in~\cite{stathopoulos2023learning}.
\end{table}

\paragraph{\methodname{} vs. SOTA with larger real dataset.}

We extend our evaluation of \dinohmr{} and \methodname{} by training them with a larger dataset from MagicPony \cite{wu2023magicpony}, denoted as \dinohmr{}$\star$ and \methodname{}$\star$, and conducting assessments using the PASCAL horse subset~\cite{everingham2015pascal}. For real images, we apply both silhouette and keypoints loss.

\begin{wraptable}{r}{5.5cm}
\centering
\captionsetup{font=normalsize}
\caption{\methodname{} v.s. SOTA methods}
\label{tab:combined}
    \begin{subtable}{\linewidth}
    \centering
\caption{Keypoint Transfer Tasks on Pascal. $\dagger$: number taken from~\cite{li2024learning,jakab2023farm3d}.}
\label{tab:comparison_magicpony}
\resizebox{0.6\linewidth}{!}{
\begin{tabular}{c|c} 
\toprule
Network & PCK\tiny{@0.1}\normalsize$\uparrow$ \\
\midrule
UMR~\cite{li2020self}$\dagger$ & 28.4  \\
A-CSM~\cite{kulkarni2020articulation}$\dagger$ & 32.9  \\ 
MagicPony~\cite{wu2023magicpony}$\dagger$ & 42.9  \\ 
Farm3D~\cite{jakab2023farm3d}$\dagger$ & 42.5  \\ 
3D Fauna~\cite{li2024learning}$\dagger$ & 53.9  \\ 
\midrule
\dinohmr{} & 33.8 \\ 
\methodname{} & 37.1 \\ 
\dinohmr{}$\star$ & 54.3 \\ 
\methodname{}$\star$ & \textbf{55.8} \\ 
\bottomrule
\end{tabular}
}
\end{subtable}

\begin{subtable}{\linewidth}
    \centering
\caption{3D evaluation on PFERD~\cite{PFERD}.}\label{tab:chamfer}
\resizebox{0.8\linewidth}{!}{
\begin{tabular}{cccc}
\toprule
Network & Dessie$\star$ & MagicPony & 3D Fauna \\
CD (mm)$\downarrow$ &  \textbf{11.85} & 32.28 & 30.10 \\
\bottomrule
\end{tabular}
}
\end{subtable}
\footnotesize{ \\ Note:  $\star$: Model fine-tuned on the whole dataset in~\cite{wu2023magicpony}.} 
\end{wraptable}
We evaluate the quality of keypoint transfer~\cite{kulkarni2020articulation}, by generating source-target image pairs from the PASCAL validation list. For each source image, we project the visible vertices of the predicted mesh onto it and align each annotated 2D ground-truth keypoint with its closest vertex. This vertex is projected onto the target image to compute the error with the ground-truth target points, quantified by PCK@0.1.
Table~\ref{tab:comparison_magicpony} shows that \methodname{} outperforms the latest learning-based SOTAs. Remarkably, 3D Fauna is trained on a dataset larger than MagicPony.

We further conduct a quantitative 3D evaluation on a subset of PFERD~\cite{PFERD} by computing the Chamfer Distance (CD) after Procrustes alignment between the 3D GT and predicted results. As shown in Table~\ref{tab:chamfer}, \methodname{} outperforms other SOTAs.

\paragraph{Qualitative and Generalization Results: \dinohmr{} vs. \methodname{}.} 

\begin{figure}[t]
    \centering
    \includegraphics[width = 0.95\linewidth]{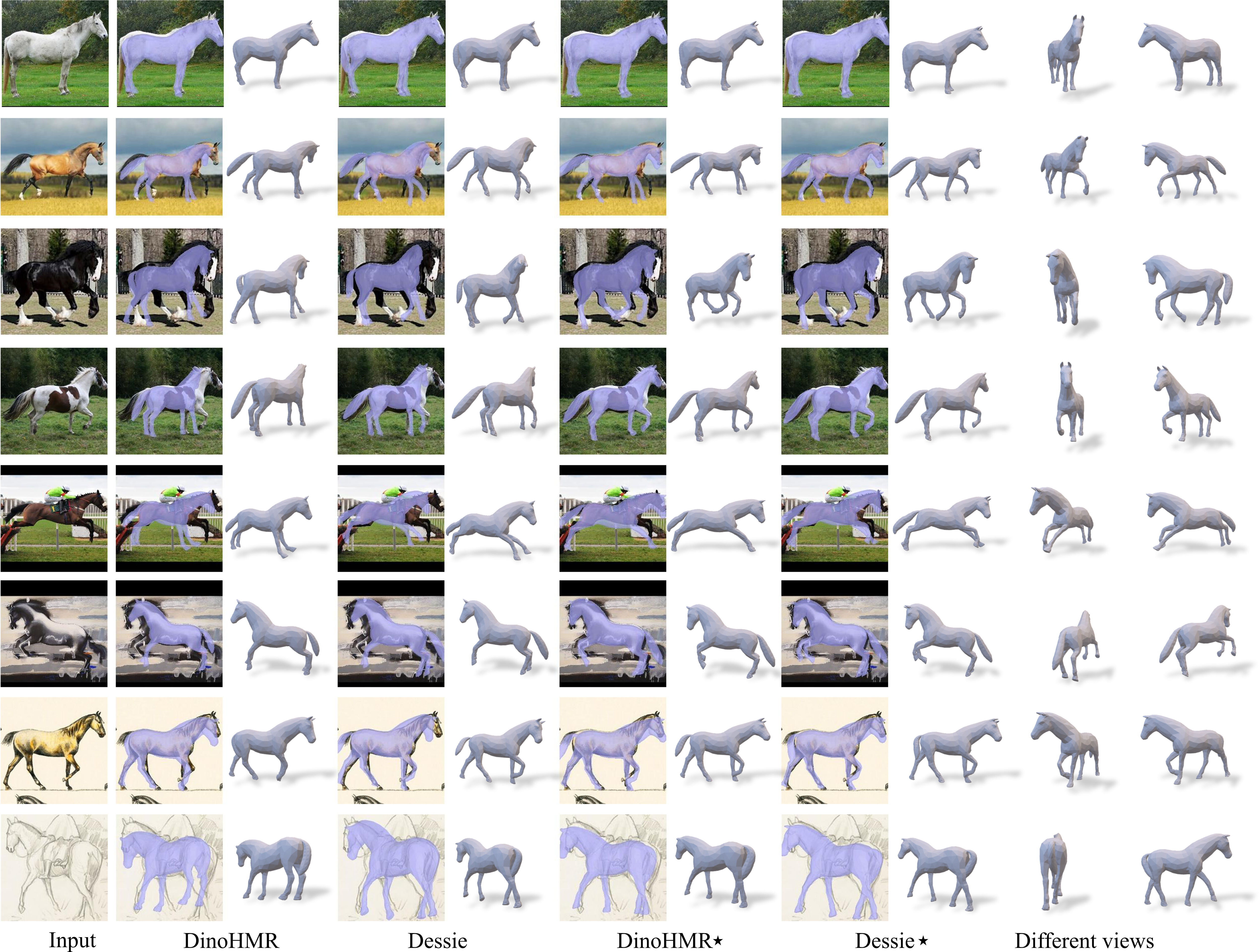}
 \caption{\dinohmr{} and \methodname{} before and after$^{\star}$ real-world data fine-tuning.}
    \label{fig:comparison_self_based}
\end{figure}

We qualitatively compare \dinohmr{} and \methodname{}, trained with and without real-world data, as depicted in Fig~\ref{fig:comparison_self_based}. Incorporating additional real-world data in training enhances both models' performance. 
Both methods exhibit robust capabilities across different image domains, even when applied to out-of-distribution data (the last three rows in Fig~\ref{fig:comparison_self_based}), highlighting their adaptability and effectiveness.

\begin{figure}[!p]
    \centering
    \includegraphics[width=0.95\linewidth]{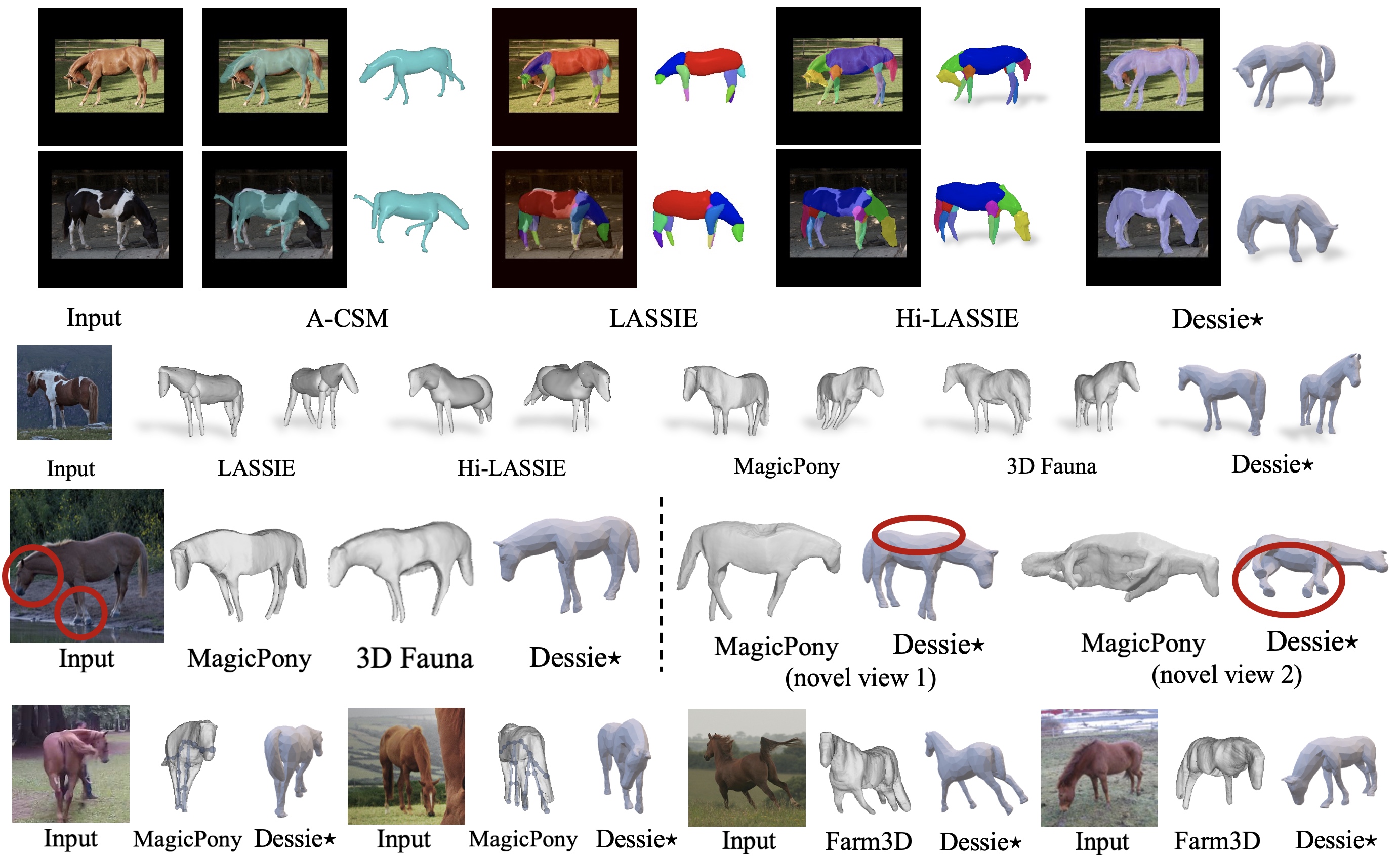}
    \caption{Qualitative comparison between SOTA and \methodname{}. Non-Dessie figures are taken from the original publications~\cite{yao2023hi,jakab2023farm3d, wu2023magicpony, li2024learning}, except for the third row.} \label{fig:qualitative_results_sota}
\end{figure}

\paragraph{Qualitative Results: \methodname{} vs. SOTA.}
Fig.~\ref{fig:qualitative_results_sota} showcases \methodname{} versus SOTA methods, highlighting better mesh reconstruction results. In addition to shape and pose reconstruction, it has better global orientation prediction in challenging scenarios where MagicPony and Farm3D struggle as shown in the last rows.

\begin{figure}[!p]
    \centering
\includegraphics[width=0.95\linewidth]{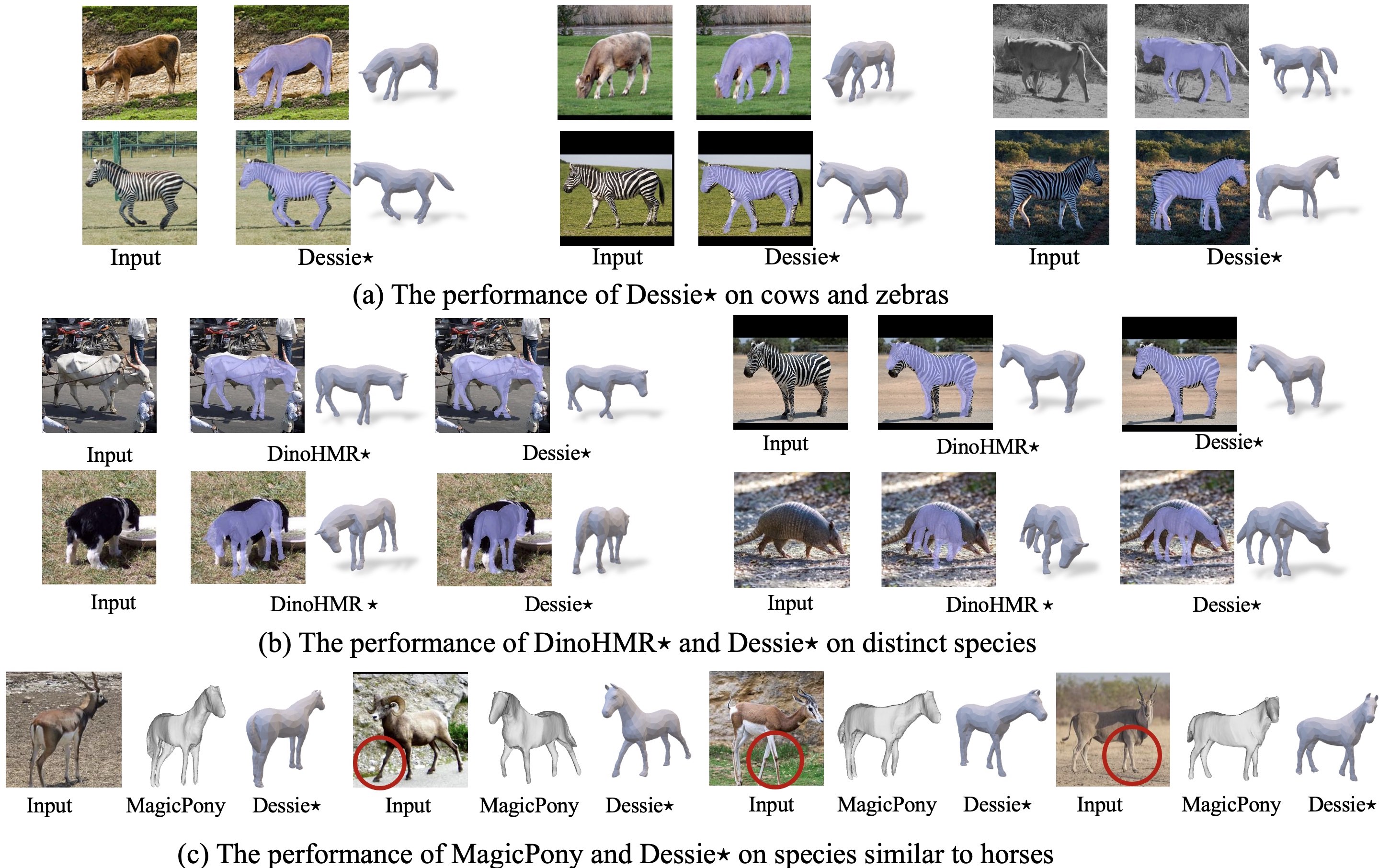}
\caption{Reconstructions of other species not present in the training set. }
\label{fig:other_species}
\end{figure}

\paragraph{Qualitative Results: \methodname{} with other species.}
Fig.~\ref{fig:other_species}(a,b) demonstrates \methodname{} reconstructing the 3D mesh better for horse-like and other distinct species. Despite being fine-tuned solely on limited horse data, \methodname{} captures better poses (Fig.~\ref{fig:other_species}(c)) compared to MagicPony, which is also trained on horse data. 
\section{Conclusion} 
\label{sec:discussion}

In this work, we introduce \dinohmr{} and \methodname{}, two simple and effective frameworks for horse mesh recovery based on the vision foundational model DINO, utilizing the synthetic data generation pipeline \pipeline{}. 
The performances further increase with finetuning using real images. Our experiments show that \methodname{} generalizes well to unseen real images, both horses with new skin patterns and shapes, but also similar species like zebra, cow and deer.
\methodname{} is a step towards a viable solution, particularly, for rare species where the scarcity of data poses a big challenge to training comprehensive models from scratch. Our results indicate that even with a minimal dataset for fine-tuning, \methodname{} maintains its efficacy, underscoring its applicability for scarce-data animal applications.

\begin{figure}[!p]
\centering
\includegraphics[width=0.8\textwidth]{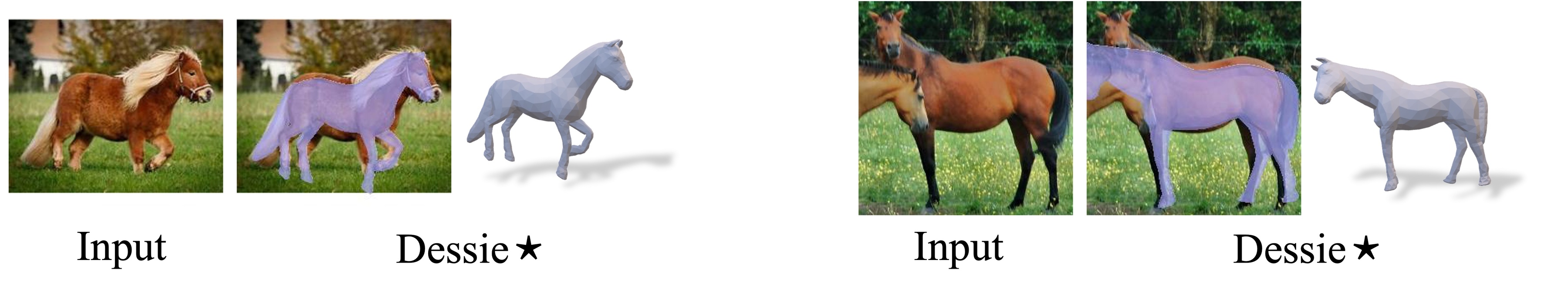}
\caption{Failure cases. Left: Novel shapes. Right: Two subjects.}
\label{fig:failure}
\end{figure}

\paragraph*{Limitations.}
Currently, \methodname{} operates on cropped images with a single subject, which poses challenges if this setting is violated (Fig.~\ref{fig:failure}, right). To overcome the limited shape expressiveness of the hSMAL model, which generalizes poorly to shapes very different from those in the training set (Fig.~\ref{fig:failure}, left), the latest horse shape model, VAREN~\cite{Zuffi:CVPR:2024} can be used.
Finally, while model-based approaches can resolve ambiguities deriving from monocular settings, they require learning a shape and pose prior, which can be challenging for the animal species of interest.

\paragraph*{\em\bf Acknowledgements.}
We thank Qingwen Zhang, Alfredo Reichlin, and Giovanni Luca Marchetti for the fruitful discussions. Silvia Zuffi is supported by PNRR FAIR Future AI Research (PE00000013), Spoke 8 Pervasive AI (CUP H97G22000210007) under the NRRP MUR program by NextGenerationEU. Elin Hernlund is supported by an SLU career grant no. 57122058, and the Beijer foundation. The computations were enabled by the supercomputing resource Berzelius provided by the National Supercomputer Centre at Linköping University and the Knut and Alice Wallenberg Foundation, Sweden.


\newpage


\section*{Supplementary material}

# Supplementary Material for:
# Dessie: Disentanglement for Articulated 3D Horse Shape and Pose Estimation from Images

Ci Li[1], Yi Yang[1,2], Zehang Weng[1],
Elin Hernlund[3], Silvia Zuffi[4], and Hedvig Kjellström[1,3]

[1] KTH, Sweden {cil, yiya, zehang, hedvig}@kth.se
[2] Scania, Sweden carol-yi.yang@scania.com
[3] SLU, Sweden Elin.Hernlund@slu.se
[4] IMATI-CNR, Italy silvia@mi.imati.cnr.it

This document provides the supplementary materials. Section 1 presents how we employ a diffusion model in DessiePIPE to synthesize realistic UV texture maps. In Section 2, we offer additional qualitative results using out-of-domain images. Section 3 offers visualization of the DINO key features for DinoHMR. Section 4 offers additional analysis regarding training Dessie with 3D GT label.

## 1 Texture Generation

We utilized the TEXTure [3] to create 80 realistic UV texture maps for our DessiePIPE. We construct specific prompts for eight unique horse species by employing the format: *A photo of a <SPECIES NAME>, {} view.*

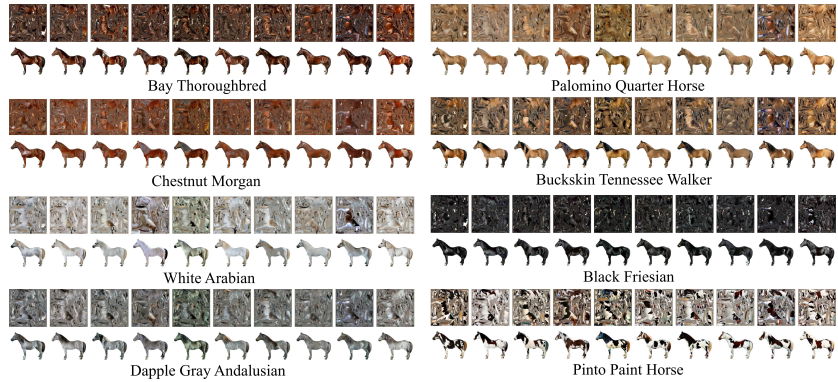

**Fig. 1:** UV texture maps created using TEXTure with eight horse species and rendered with the hSMAL model in zero pose.

<SPECIES NAME> is substituted with specific names, including "Bay Thoroughbred", "Palomino Quarter Horse", "Chestnut Morgan", "Buckskin Tennessee Walker", "White Arabian", "Black Friesian", "Dapple Gray Andalusian", and "Pinto Paint Horse". For each prompt, {}*view* is filled in with "front", "left",

"back", "right", "overhead" and "bottom", enabling the creation of textures for the horse model from various viewpoints through rotation. For each species, ten texture maps are generated using ten different random seeds. The resulting texture maps are shown in Fig. 1.

## 2    Qualitative Results

We show more reconstruction results of images from different domains in Fig. 2.

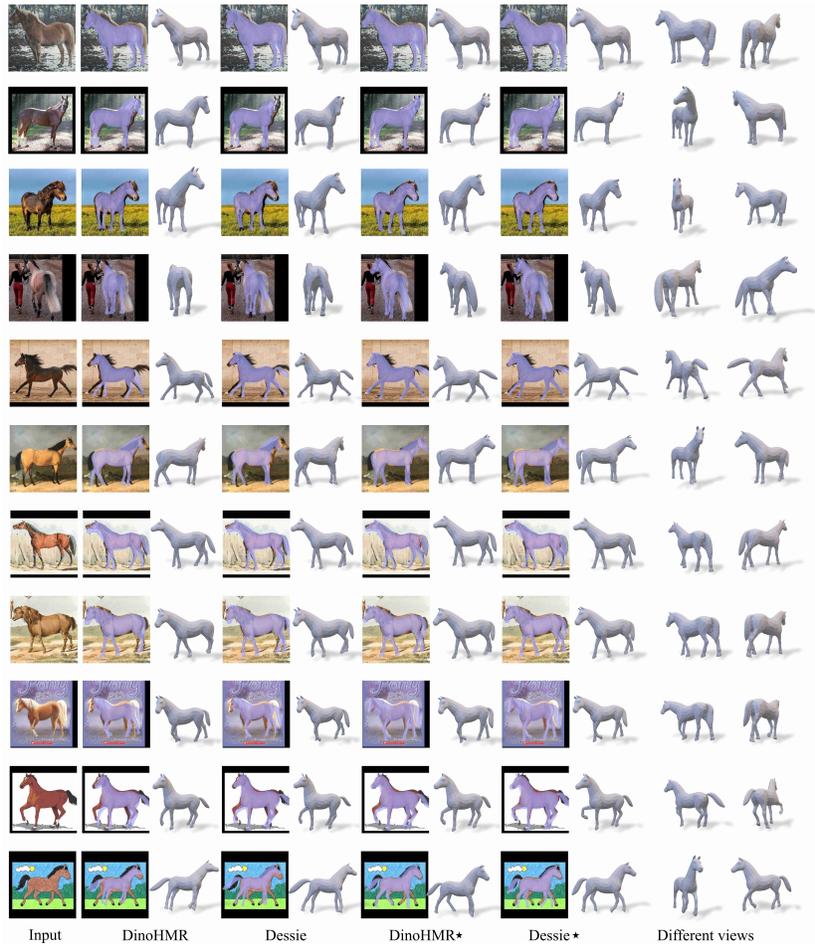

**Fig. 2:** More qualitative results of DinoHMR and Dessie before and after fine-tuning with real-world data. The first four rows showcase results from real-world images, while the subsequent rows are for images of horses from various domains, such as oil paintings and cartoons.

# 3 DINO Key Feature

The DINO features in the DinoHMR, visualised in the same way we mention in the main paper (Section 4.3), also focus on the subject.

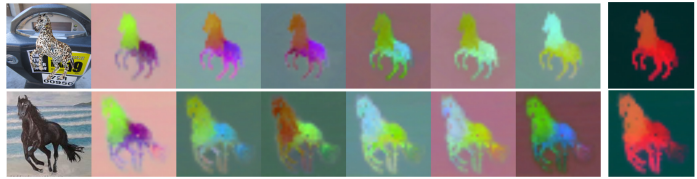

**Fig. 3:** Visualization of the leading PCA components of DinoHMR key features.

# 4 GT Label Effect

We investigate the effect of incorporating 3D GT labels in training Dessie with DessiePIPE. More specifically, we evaluate the skeletal joints in five real-horse sequences from PFERD [2], using 3D metric PAMPJPE and 2D metric PCK@0.1. For each skeletal joint, we manually define the correspondence on the mesh. For the 3D metric, we compare the predicted results with the 3D hSMAL GT, and for the 2D metric, we project the mesh vertices and compare them with the 2D GT. The results in Table 1 show that 3D GT loss $L_{gt}$ improves 3D but degrades 2D performance. This is potentially caused by the predicted weak perspective camera, which does not match the actual camera, resulting in a mismatch between the projected 3D points and the detected 2D ones, as noted in TokenHMR [1].

**Table 1:** 3D evaluation on PFERD for Dessie.

| $L_{gt}$ | PAMPJPE ↓ | PCK@0.1 ↑ |
|---|---|---|
| ✓ | **102.59** | 0.90 |
| - | 119.36 | **0.92** |

✓ or - indicates inclusion or exclusion of the loss component in training.


\end{document}